\pdfoutput=1
\documentclass{article} 
\usepackage{nips14submit_e,times}
\usepackage{url}
\usepackage{amsmath,amsfonts,graphicx}

\def\bx{{\mbox{\boldmath $x$}}}
\def\bw{{\mbox{\boldmath $w$}}}
\newcommand{\vect}[1]{\boldsymbol{#1}}

\title{Autoencoder Trees}

\author{
Ozan \.Irsoy\\
Department of Computer Science\\
Cornell University\\
Ithaca, NY 14853 USA \\
\texttt{oirsoy@cs.cornell.edu} \\
\And
Ethem Alpayd{\i}n \\
Department of Computer Engineering \\
Bo\u{g}azi\c{c}i University \\
Bebek, \.Istanbul 34342 Turkey \\
\texttt{alpaydin@boun.edu.tr} \\
}

\nipsfinalcopy 

\begin{document}

\maketitle

\begin{abstract}
We discuss an autoencoder model in which the encoding and decoding functions are implemented by decision trees. We use the soft decision tree where internal nodes realize soft multivariate splits given by a gating function and the overall output is the average of all leaves weighted by the gating values on their path. The encoder tree takes the input and generates a lower dimensional representation in the leaves and the decoder tree takes this and reconstructs the original input. Exploiting the continuity of the trees, autoencoder trees are trained with stochastic gradient descent. On handwritten digit and news data, we see that the autoencoder trees yield good reconstruction error compared to traditional autoencoder perceptrons. We also see that the autoencoder tree captures hierarchical representations at different granularities of the data on its different levels and the leaves capture the localities in the input space.
\end{abstract}

\section{Introduction}

To find the hidden structure in data, one unsupervised learning method is the autoencoder which is composed of an encoder and a decoder put back to back. The encoder maps the original input to a generally lower dimensional or sparse hidden representation, and the decoder takes it and reconstructs the original input. The idea is that if the decoder can reconstruct the original input faithfully, the hidden representation should be a meaningful and useful one. Conventional autoencoders use use a single layer, perceptron-type neural network for the encoding and decoding functions, which implements an affine map followed by a nonlinearity for the encoder~\cite{cottrell1987learning}. In deep learning, the idea is to stack multiple such autoencoders as a multilayer perceptron thereby learning more abstract hidden representations~\cite{bengio2009learning}.

In this work, we explore an autoencoder model where decision trees are used for the encoding and decoding functions, instead of a single or multilayer perceptrons. We use the soft decision tree model where the internal decision nodes use a soft multivariate split defined by a gating function and the overall output is the average of all leaves weighted by the gating values on their paths (section~\ref{sec-sdt}). Since the output of such a soft tree is continuous, we can use stochastic gradient-descent to update the parameters of encoder and decoder trees simultaneously to minimize reconstruction error, as in conventional autoencoder approaches. With the chain rule, error terms of the hidden representation of the decoding layer are passed back to the encoding layer as its external error (section~\ref{sec-auto}). We will show in our experimental results (section~\ref{sec-exp}) that such autoencoder trees can learn as well as autoencoder perceptrons while learning a hierarchical decomposition of the data into subspaces which respect localities in the data.

\section{Soft Decision Tree}
\label{sec-sdt}

A decision tree is a hierarchical structure with internal decision nodes and terminal leaves. An internal decision node redirects the given instance to one of its children. Leaves traditionally include a prediction label, such as a class label for classification or a numeric response for regression.

As opposed to the hard decision node which implements a hard split, a soft decision node redirects instances to all its children but with different probabilities, as given by a {\em gating function\/} $g_m(\bx)$---the hard decision tree is a special case where $g(\bx)\in\{0,1\}$~\cite{irsoy2012soft}. Let us consider a {\em soft binary tree\/} where each internal node has two children, named left and right. The response at a node is recursively calculated as the weighted average of the responses of its left and right children: 
\begin{align}
y_m(\vect{x}) =
  \begin{cases} 
      \hfill \rho_m    \hfill & \text{if $m$ is leaf} \\
      \hfill g_m(\vect{x}) y_{ml}(\vect{x}) + (1-g_m(\vect{x})) y_{mr}(\vect{x}) 
              \hfill & \text{otherwise} \label{eq:split}\\
  \end{cases}
\end{align}	

To choose among the two outcomes, we define the gating function, $g_m(\bx)\in [0,1]$, as the {\em sigmoid function\/} over a linear split: 
\begin{align}
g_m(\bx) = \frac{1}{1+\exp[-(\bw_m^T\bx)]} \label{eq:gating}
\end{align}

Note that our decision nodes are {\em multivariate}, that is, they use all of the input features, as opposed to the univariate trees that use a single feature in each split. Geometrically speaking, though univariate splits are  orthogonal to one of the axes, multivariate splits are {\em oblique\/} and can take any orientation, which makes them more generally applicable.

Separating the regions of responsibility of the left and right children can be seen as two-class classification problem and from that perspective, the gating model implements a discriminative (logistic linear) model estimating the posterior probability of the left child: $P(\mbox{left}|\bx)\equiv g_m(\bx)$ and $P(\mbox{right}|\bx)\equiv 1-g_m(\bx)$. This architecture is equivalent to that of the hierarchical mixture of experts~\cite{jordan1994hierarchical}.

In the case of supervised learning, $\rho$ stored at a leaf corresponds to the predicted value. In regression, $\rho$ is a scalar. In classification, we can apply a sigmoid nonlinearity at the root node to convert the output into a probability value. Note that $\vect{\rho}$ can be a vector response as well: For example with $K\ge 2$ classes, $\vect{\rho}$ is a $K$-dimensional vector and softmax nonlinearity is used at the root node to convert the $K$ outputs to posterior probabilities. In our case of unsupervised learning, the dimensionality of $\vect{\rho}$ will be set to the dimensionality of the hidden representation we want to learn.

Note that the soft decision tree defines a continuous response function of the parameter space, conditioned on the structure of the tree. This means that given a tree structure, the parameters, $\{\vect{\rho}_m, \vect{w}_m\}_m$(response values at the leaves and splitting hyperplanes of the internal nodes) can be learned by minimizing an objective function over the tree response with a continuous optimization method, e.g., stochastic gradient-descent. For supervised tasks such as classification or regression, conventional objective functions such as cross-entropy or squared error can be employed. 

In order to learn the parameters $\vect{\rho}$ and $\vect{w}$ with a gradient-based method, we can use backpropagation to efficiently compute the gradients. Let us define $\vect{\delta}_m = \partial E^i/\partial \vect{y}_m(\vect{x}^i)$, which is the \emph{responsibility\/} of the node $m$. Backpropagating the error from the root towards the leaves, we have
\begin{align}
\dfrac{\partial E^i}{\partial \vect{w}_{m}} &= g_m(\vect{x}^i) (1-g_m(\vect{x}^i)) \big(\vect{\delta}^{iT}_m(\vect{y}_{ml}(\vect{x}^i) - \vect{y}_{mr}(\vect{x}^i))\big)\vect{x}^i\\
\dfrac{\partial E^i}{\partial \vect{\rho}_{m}} &= \vect{\delta}^i_m
\end{align}
with 
\begin{align}
 \vect{\delta}^i_m =
  \begin{cases} 
      \hfill \vect{y}_r(\vect{x}) - \vect{r}^i \hfill & \text{ if $m$ is root}\\
      \hfill \vect{\delta}^i_{pa(m)}g_m(\vect{x}^i)   
              \hfill & \text{ if $m$ is a left child} \\
      \hfill \vect{\delta}^i_{pa(m)}(1-g_m(\vect{x}^i))   
              \hfill & \text{ if $m$ is a right child} \\
  \end{cases}
\end{align}

where $pa(m)$ is the parent of node $m$.

\section{Autoencoder Trees}
\label{sec-auto}

\begin{figure}[t]
\centering
\includegraphics[scale=.65]{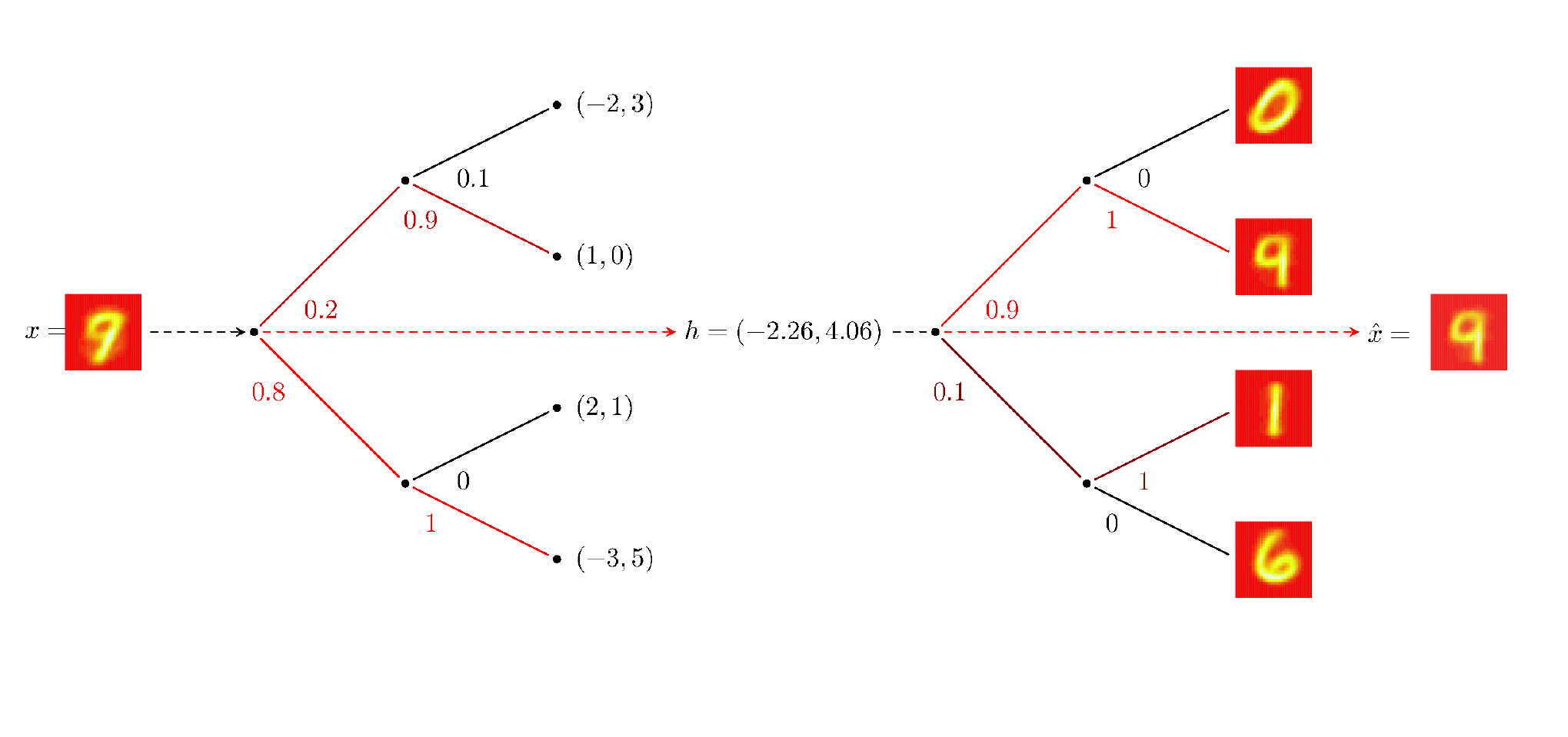}
\caption{Operation of an autoencoder tree. Given a 784d input image $x$, the encoder tree assigns it a 2d hidden representation $h=(-2.26,4.06)$ by making a soft selection among its leaf responses (left). Similarly, decoder tree reconstructs the input when given this $h$ (right).}
\label{fig:tree0}
\end{figure}

The observation that the soft decision tree output is a a continuous function of the parameter space (for a given tree structure) leads to the following conclusion: A soft decision tree can be trained not just with supervised error signal but also with an unsupervised signal, as well as an error term backpropagated from a further layer of information processing which uses the tree output as an input. This follows from a simple application of the chain rule when computing the gradients.

In this work, based on this observation, we define an autoencoder approach by stacking two soft trees back to back where the encoding and decoding functions are implemented by two soft decision trees. 

Let $\vect{t} (\vect{x})$ denote a soft decision tree as defined in Equations \ref{eq:split} and \ref{eq:gating}. Let us define an autoencoder tree pair $\vect{t}_e (\vect{x})$ and $\vect{t}_d (\vect{x})$ with the following interpretation: The encoder tree encodes the $d$ dimensional input $\vect{x}$ into a
$k$ dimensional intermediate (or hidden) representation $\vect{h} = \vect{t}_e(\vect{x})$ (where $k<d$), and the decoder tree decodes the initial input from the hidden representation, $\hat{\vect{x}} = \vect{t}_d (\vect{h})$ (Figure~\ref{fig:tree0}). We want the decoded response to be as close as possible to the initial input ($\vect{x} \approx \hat{\vect{x}}$)
which can be implemented by minimizing the reconstruction error on a training set $\{\vect{x}^i\}_{i=1}^N$:
\begin{equation}
E = \frac{1}{2}\sum_i \|\vect{x}^i - \hat{\vect{x}}^i\|^2 = \frac{1}{2}\sum_i \|\vect{x}^i - \vect{t}_d(\vect{t}_e(\vect{x}^i))\|^2
\end{equation}

Let $\theta_e \in \{\vect{\rho}_m, \vect{w}_m\}_{m \in e}$ and $\theta_d \in \{\vect{\rho}_m, \vect{w}_m\}_{m \in d}$ be a parameter of the encoder and decoder trees respectively. Then, we can update parameters in both trees by gradient descent:
\begin{align}
\dfrac{\partial E^i}{\partial \theta_d} = \dfrac{\partial E^i}{\partial \vect{t}_d(\vect{h}^i)}\dfrac{\partial \vect{t}_d(\vect{h}^i)}{\partial \theta_d}
 = (\vect{x}^i - \hat{\vect{x}}^i)^T\dfrac{\partial \vect{t}_d(\vect{h}^i)}{\partial \theta_d}\\
 \dfrac{\partial E^i}{\partial \theta_e} = \dfrac{\partial E^i}{\partial \vect{h}^i}\dfrac{\partial \vect{h}^i}{\partial \theta_e}
 = (\vect{x}^i - \hat{\vect{x}}^i)^T\dfrac{\partial \vect{t}_e(\vect{h}^i)}{\partial \vect{h}^i}\dfrac{\partial \vect{t}_e(\vect{x}^i)}{\partial \theta_e}
\end{align}
where $\dfrac{\partial \vect{t}_{\cdot}(\vect{x})}{\partial \theta_{\cdot}}$ can be computed as before. Additionally, the computation of
$\vect{\delta h} = \dfrac{\partial E}{\partial \vect{h}}$ requires the derivative of a tree response with respect to its input (for the decoder tree):
\begin{align}
\dfrac{\partial \vect{t}(\vect{x}^i)}{\partial \vect{x}^i} &= \sum_m g_m(\vect{x}^i) (1-g_m(\vect{x}^i)) 
                 \big(\delta^{iT}_m (y_{ml}(\vect{x}^i) - y_{mr}(\vect{x}^i))\big)w
\end{align}
That is, $\delta h$ is the error responsibility of the hidden representation backpropagated from the decoder tree (top layer) to the encoder tree (bottom tree).

When the encoder and decoder trees have multiple levels, backpropagating may be too slow and we use a layer-by-layer training, as in conventional autoencoder training. For both the encoder and the decoder trees, we start with a tree of depth two. After iterating for some number of epochs, we split every leaf into a tree and hence get trees of depth three, and we continue doing so until we get to the final required depth. During those depth increments, all tree parameters are updated and not just the most recently introduced ones, which allows for finetuning. When splitting, the new children inherit the response $\rho$ values of their parents with an additive small random noise.

\section{Experiments}
\label{sec-exp}

\subsection{Experimental Setting}

\paragraph{Data.} We evaluate our models on two data sets: MNIST handwritten digit database \cite{lecun1998mnist} and the 20 Newsgroups data set~\cite{twnews}. MNIST contains 60,000 training and 10,000 test examples of handwritten digit images which are 28 by 28 pixels (784 dimensional). Output labels are the ten digits. 20 Newsgroups data (20News) contains 18,846 instances of newsgroup documents (partitioned into training and test sets with 60\%-40\% ratio), with output labels denoting the subject matter (category) out of 20 classes. With the bag-of-words representation, it has a dimensionality about 60,000; we sorted the words in terms of their frequencies, discarded the top 100 (non informative stop words) and used the next 2,000.

\paragraph{Baselines.} We use two autoencoder perceptrons: One has a single layer, that is, a linear map and nonlinearity for the encoder, and linear map for the decoder. The second uses the stacked two-layer perceptron autoencoder where we first reduce the dimensionality to 50 using the conventional autoencoder, and using the 50 dimensional representation, we once again reduce to the final dimensionality. In both cases, nonlinearity is the hyperbolic tangent.

\paragraph{Tree and network training.} Both autoencoder perceptrons and autoencoder trees are trained with stochastic gradient-descent, in the online setting (i.e. minibatch size is 1). For both, we employ a diagonal variant of AdaGrad \cite{duchi2011adaptive}, which yields smooth and fast convergence. We train for a total of 240 epochs. Autoencoder trees (both the encoder and the decoder trees) start from a depth of two, and the depth is incremented at every 40th epoch, until they reach their final depth (five or six). We employ a simple L2 regularization on connection weights for autoencoder perceptrons and hyperplane split parameters ($\vect{w}_m$) and the leaf responses ($\vect{\rho}_m$) for autoencoder trees.

\subsection{Results}

\begin{figure}[t]
\includegraphics[width=0.49\textwidth]{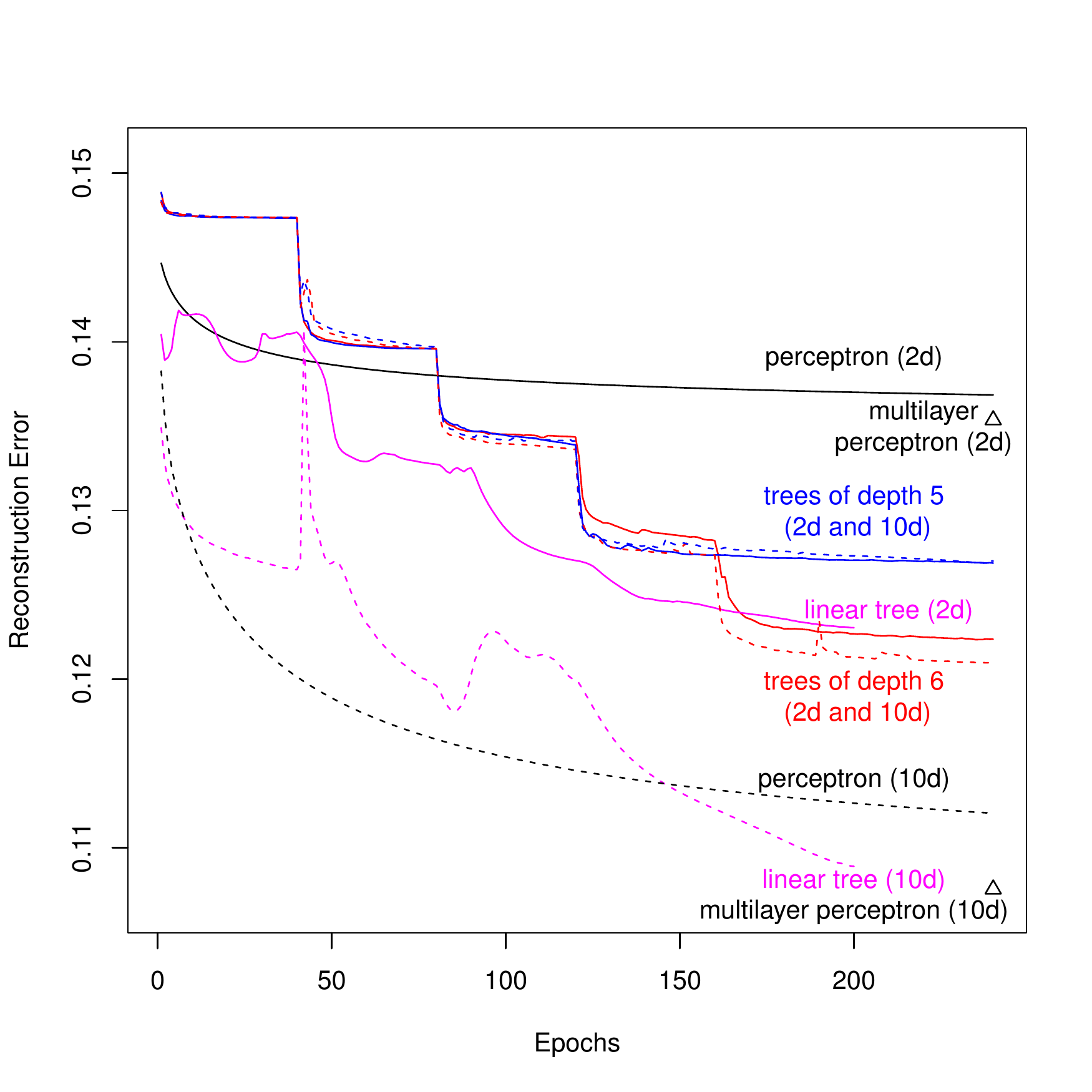}
\includegraphics[width=0.49\textwidth]{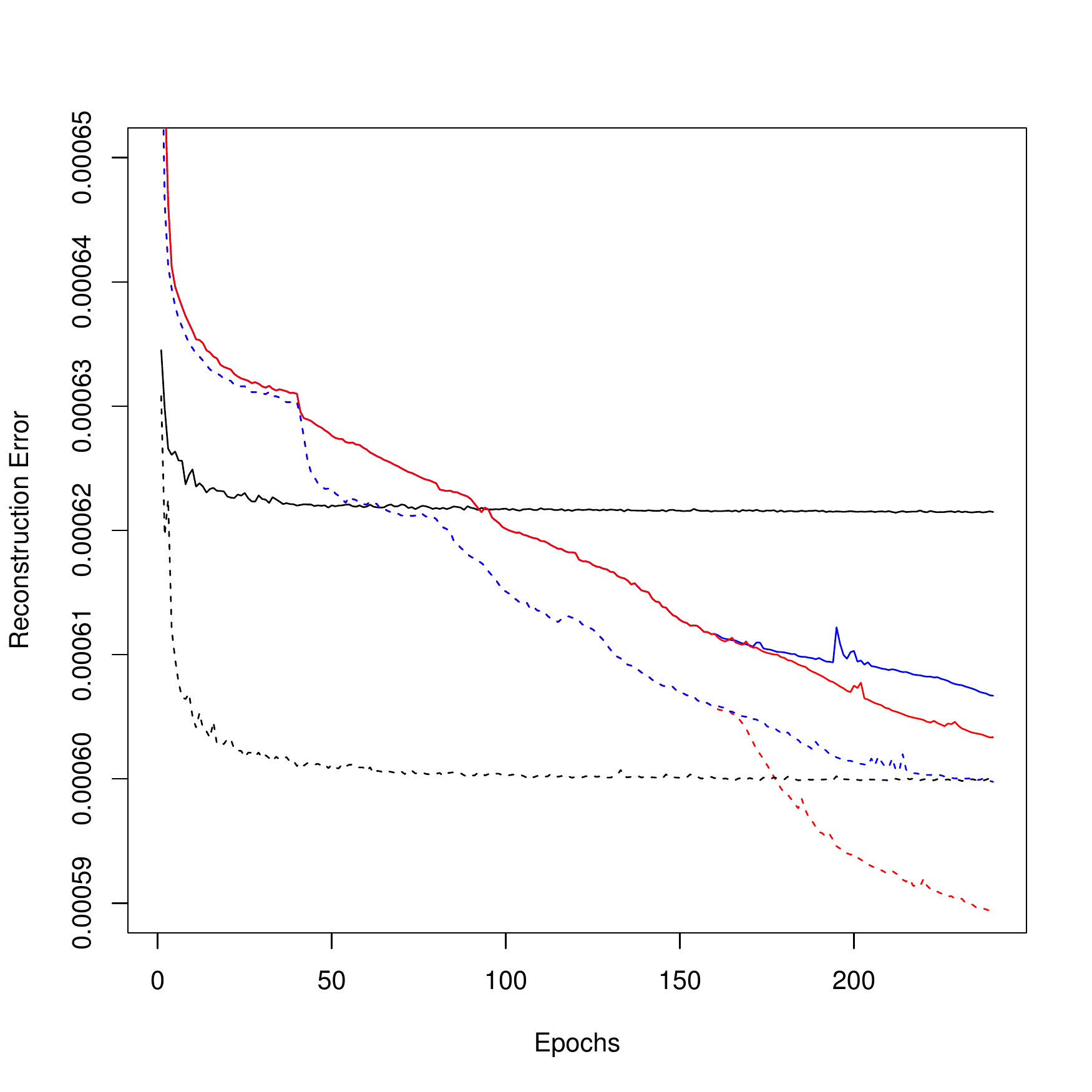}
\caption{Reconstruction errors of different autoencoder architectures on MNIST (left) and 20News (right) datasets. Black, blue and red denote the autoencoder perceptron, the autoencoder tree with a depth of five, and the autoencoder tree with a depth of six, respectively. 
Purple denote autoencoder tree with linear map at the leaves (section 4.3).
Solid and dash denote dimensionality reduction to two and ten, respectively. Points denoted by triangle show multilayer perceptron which first reduces the dimensionality to 50 and then reduces once more to 2 or 10.}
\label{fig:recerrs}
\end{figure}

We report the reconstruction errors per each epoch of stochastic gradient-descent on the two datasets, in Figure~\ref{fig:recerrs}. For MNIST, we report the error in the scale of a single pixel; for 20News, error is in the scale of a single word in the bag-of-words representation, relative to the maximum number of occurrence of each word.

We see that on MNIST, autoencoder trees can attain a better reconstruction error when reducing to two dimensions, however autoencoder perceptron is better when reducing to ten dimensions. For the autoencoder trees, we observe that the dimensionality of the hidden representation does not have a strong effect on the performance, as seen by very close reconstruction error rates. On the other hand, we observe improved performance as we increase the depth of trees. This suggests that the topology itself might be more important than the dimensionality of the hidden representation for autoencoder trees.

For 20News, convergence is less smooth for all architectures. For both reduction to two and ten, autoencoder trees yield a better reconstruction error than autoencoder perceptrons. There is a gain by reducing to ten instead of two for autoencoder trees. However, again, this gain is relatively small compared to the gain resulting from an extra level of depth in the tree, as seen by the difference between five- and six-deep trees (which is especially large in the case of reducing to ten dimensions).

\begin{figure}[t]
\centering
\includegraphics[width=0.32\textwidth]{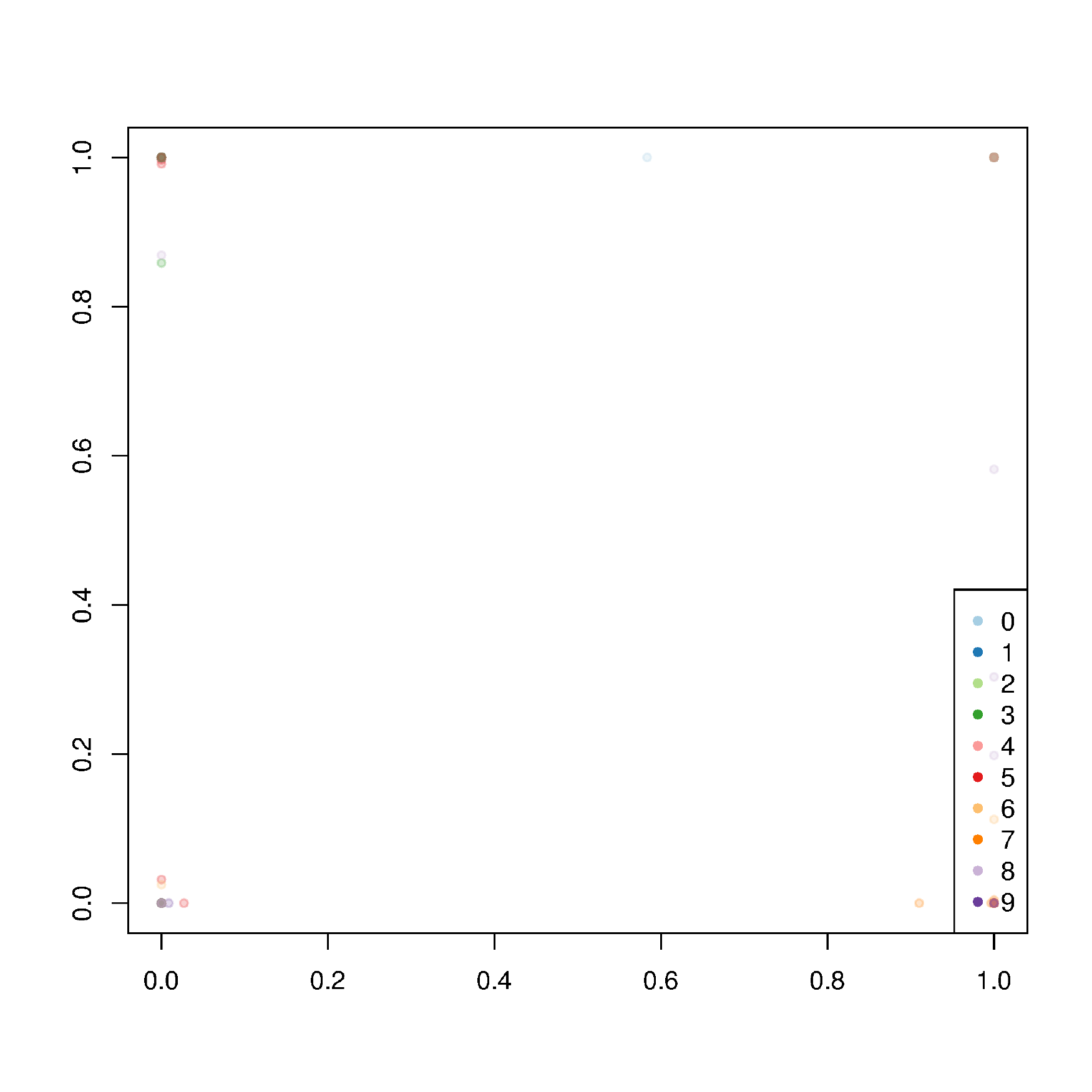}
\includegraphics[width=0.32\textwidth]{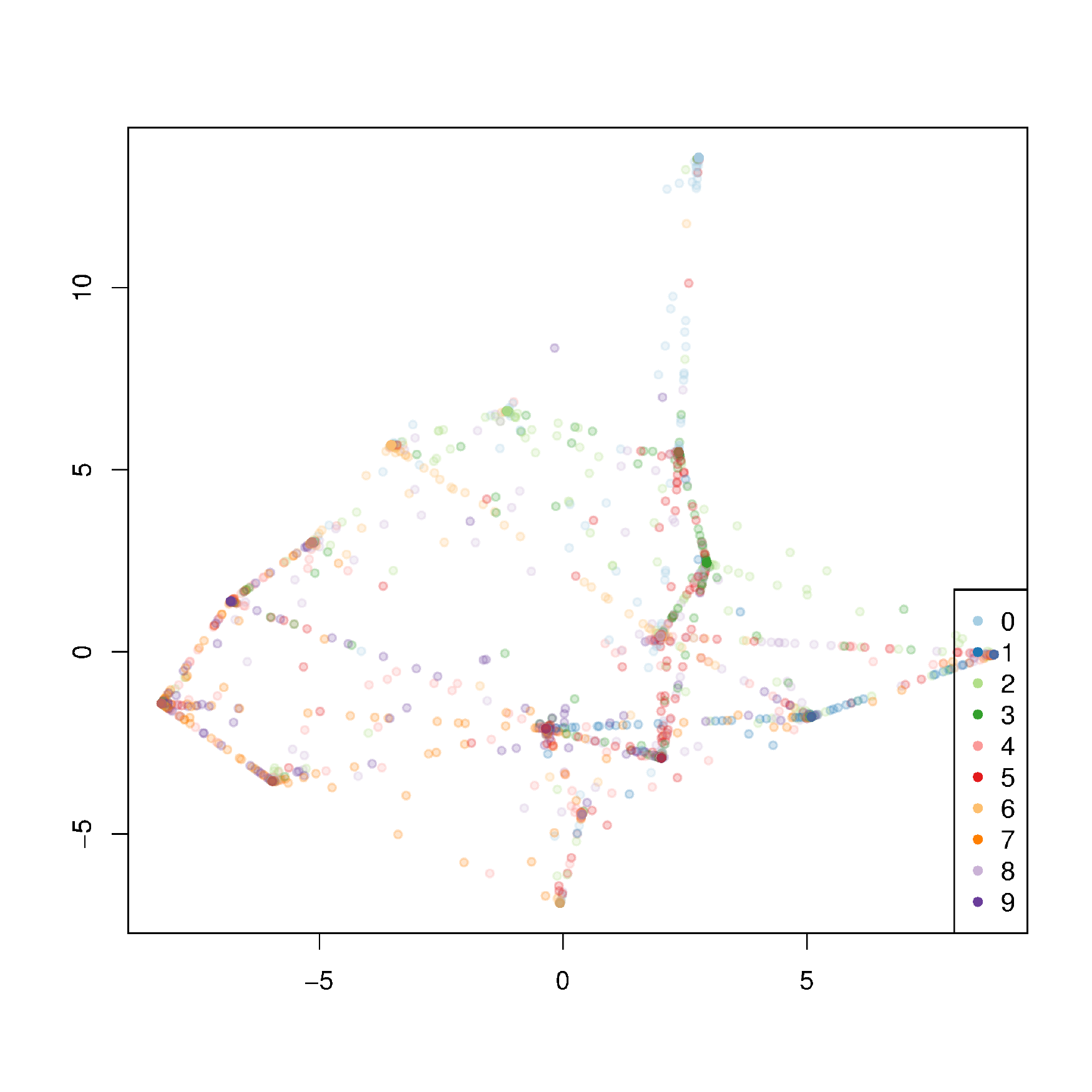}
\includegraphics[width=0.32\textwidth]{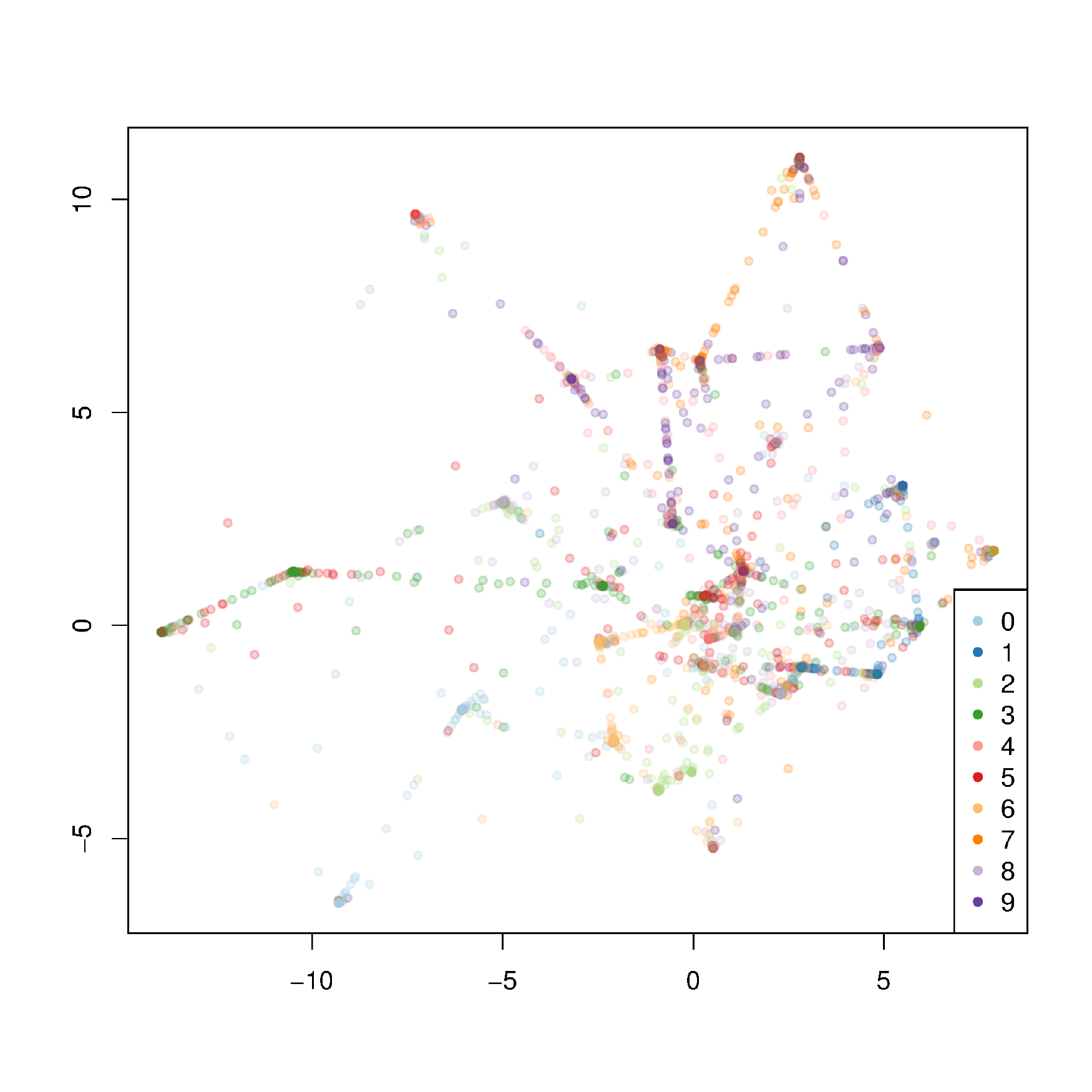}
\caption{Dimensionality reduced representations of digits for MNIST for autoencoder perceptron (left), autoencoder tree of depth five (middle) and depth six (right).}
\label{fig:dimreduced}
\end{figure}

For mapping of MNIST digits to two dimensions, we show resulting representations in Figure~\ref{fig:dimreduced}. For the autoencoder perceptron, we observe a strong tendency to saturate the nonlinearity and not utilize the softness of the sigmoid threshold and hence we see all instances mapped to the four corners. For autoencoder trees, most instances converge on a single leaf response but we also observe convex combinations of multiple leaves. Note that the hidden representations are more local rather than distributed for trees: Rather than assigning a global meaning to different directions in the hidden space, the hidden representation assigns regions (of different sizes in different levels, in a multi-resolution or multi-granular fashion) of the space to different digits, and closeness becomes more important. This behavior is similar to clustering. Indeed, autoencoder trees can be considered to do a dimensionality reduction alongside hierarchical soft clustering. Aforementioned relatively small gains by increasing the dimensionality of the hidden layer is another evidence to this local behavior, since distributed representations gain much more from a higher number of dimensions.

\begin{figure}[t]
\includegraphics[width=\textwidth]{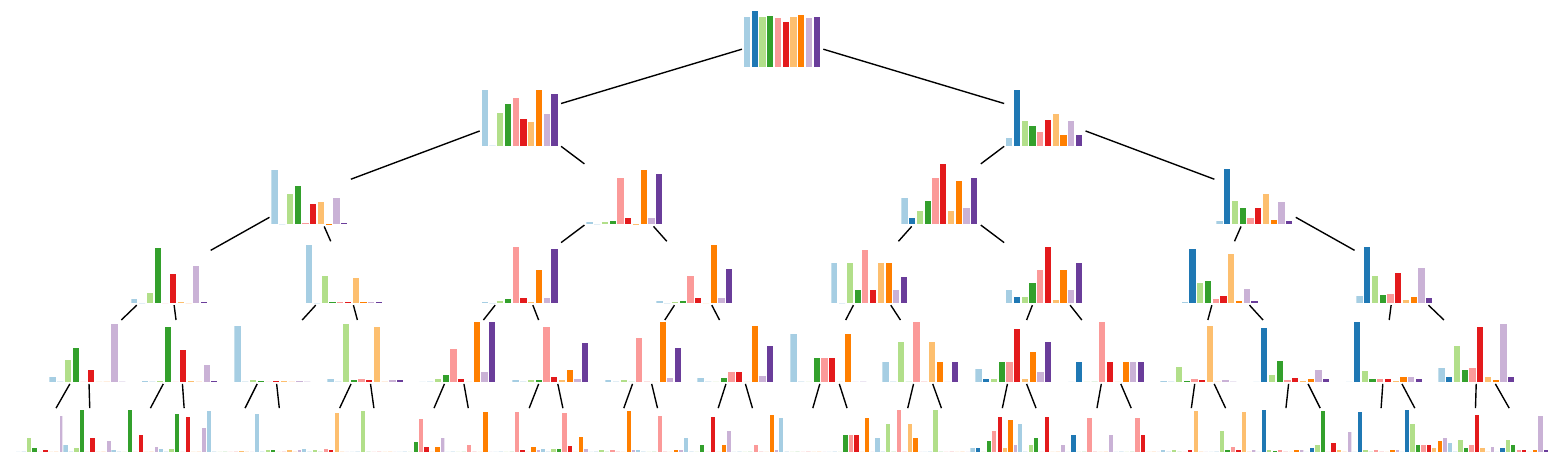}
\caption{Class distributions for the encoder tree on MNIST. Different colors represent the ten classes.}
\label{fig:mnist-encoder}
\end{figure}

In Figure~\ref{fig:mnist-encoder}, an encoder tree with a depth of six is shown. Histograms at each node shows the class distributions. Since we use a soft decision tree, every instance has a \emph{soft membership\/} at a given node, computed by the sigmoidal gating function, which is used as the soft count when counting the instances which belong to a node. Although training is unsupervised, we see that some leaf nodes  capture single classes, such as the sixth leaf including mostly only the digit `0' (light blue) or the eight leaf including only the digit `2' (light green). Others capture two, or more classes but learn a locality and the combination of gating functions from the root to that leaf defines the regions of that locality.

\begin{figure}[t]
\includegraphics[width=\textwidth]{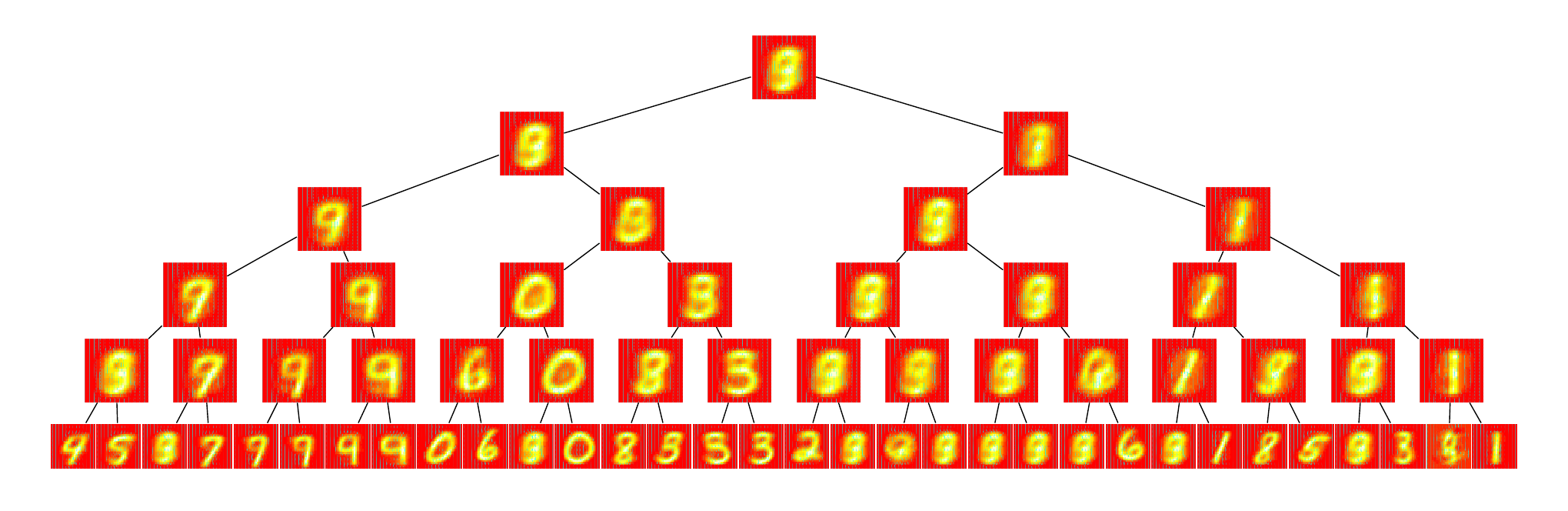}
\caption{Response values for the decoder tree on MNIST. Internal nodes show the latest response values before splitting them.}
\label{fig:mnist-decoder}
\end{figure}

In Figure~\ref{fig:mnist-decoder}, we show a decoder tree learned over the MNIST dataset. Nodes show their response values with internal nodes depicting the latest value before they are split and another level is added. We observe hierarchies captured by the decoder tree: To the left, nodes disentangle different variations of digits `9' and `7' with different slopes. At certain nodes, digits `0' and `6' are represented together, then separated at children. A similar phenomenon occurs with digits `3' and `8', as well as `2' and `8'.

\begin{figure}[t]
\centering
\includegraphics[width=0.45\textwidth]{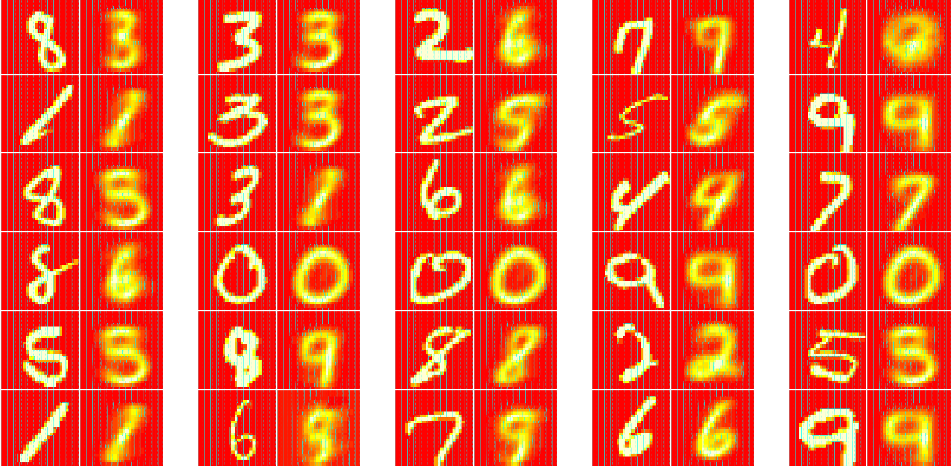} \quad \quad
\includegraphics[width=0.45\textwidth]{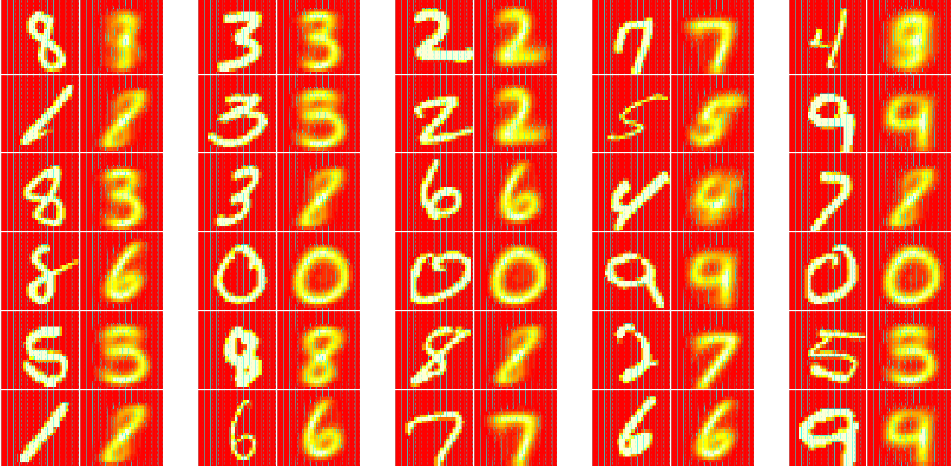}
\caption{A sample of original (left columns) and reconstructed (right columns) images using autoencoder trees. Selection is random. Left: Reconstructed from 2d reduction. Right: Reconstructed from 10d reduction.}
\label{fig:mnist-reconstr}
\end{figure}

We show some examples of original and reconstructed digit images in Figure~\ref{fig:mnist-reconstr}, for autoencoder trees with depth six and hidden dimensionality of two and ten. We observe that most reconstructions are faithful, and we see that some of the errors done by 2d reducing tree are corrected by the 10d reducing tree.

\begin{figure}[t]
\centering
\includegraphics[width=0.65\textwidth]{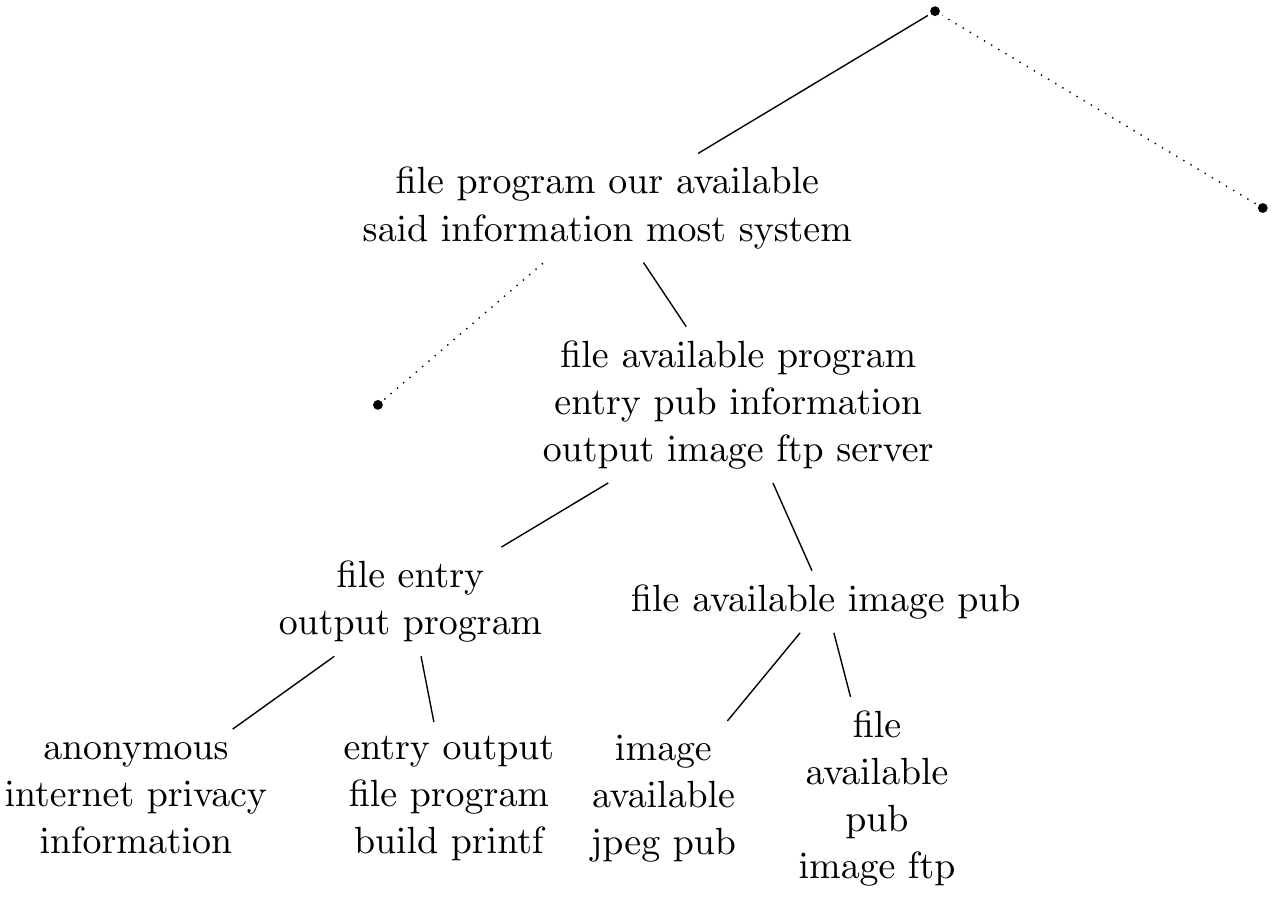}
\caption{Response values for a subtree of the decoder tree on 20News. Internal nodes show latest response values before splitting them.}
\label{fig:20news-decoder}
\end{figure}

Similarly in Figure~\ref{fig:20news-decoder}, we show part of the decoder tree over 20News dataset. Since the response vector is a bag-of-words representations, we sort the words by their coefficients and show only the top words. We show some of the paths and omit others to avoid clutter. Again, we see hierarchies captured by the tree as seen by word distributions which resemble topics at finer and finer grain as we split the nodes further. This behavior is similar to a hierarchical topic model, since every document can be mapped to a distribution on tree leaves (with the gating function) and every leaf can be mapped to a distribution on words (by normalizing leaf responses).

\subsection{Extension to Model Trees}

A simple extension to the aforementioned autoencoder tree model can be done by more complex leaf models, such as a linear map over the inputs, instead of a constant vector valued response. This can be done by modifying Equation~\ref{eq:split} so that in a leaf node $m$, we define $\vect{\rho}_m=\vect{V}_m \vect{x}$ and gradient-descent rules are modified accordingly to update $\vect{V}_m$. Hence, the value stored in a leaf is no longer constant but parameters of a linear model and the response in a leaf varies linearly based on the input.

This results in an autoencoder tree model in which leaf nodes make local linear projections in the input space. This also provides some degree of distributed representational power to the autoencoder tree, corresponding to locally partitioning the space and assigning a distributed model to every (soft) partition.

\begin{figure}[t]
\centering
\includegraphics[width=0.4\textwidth]{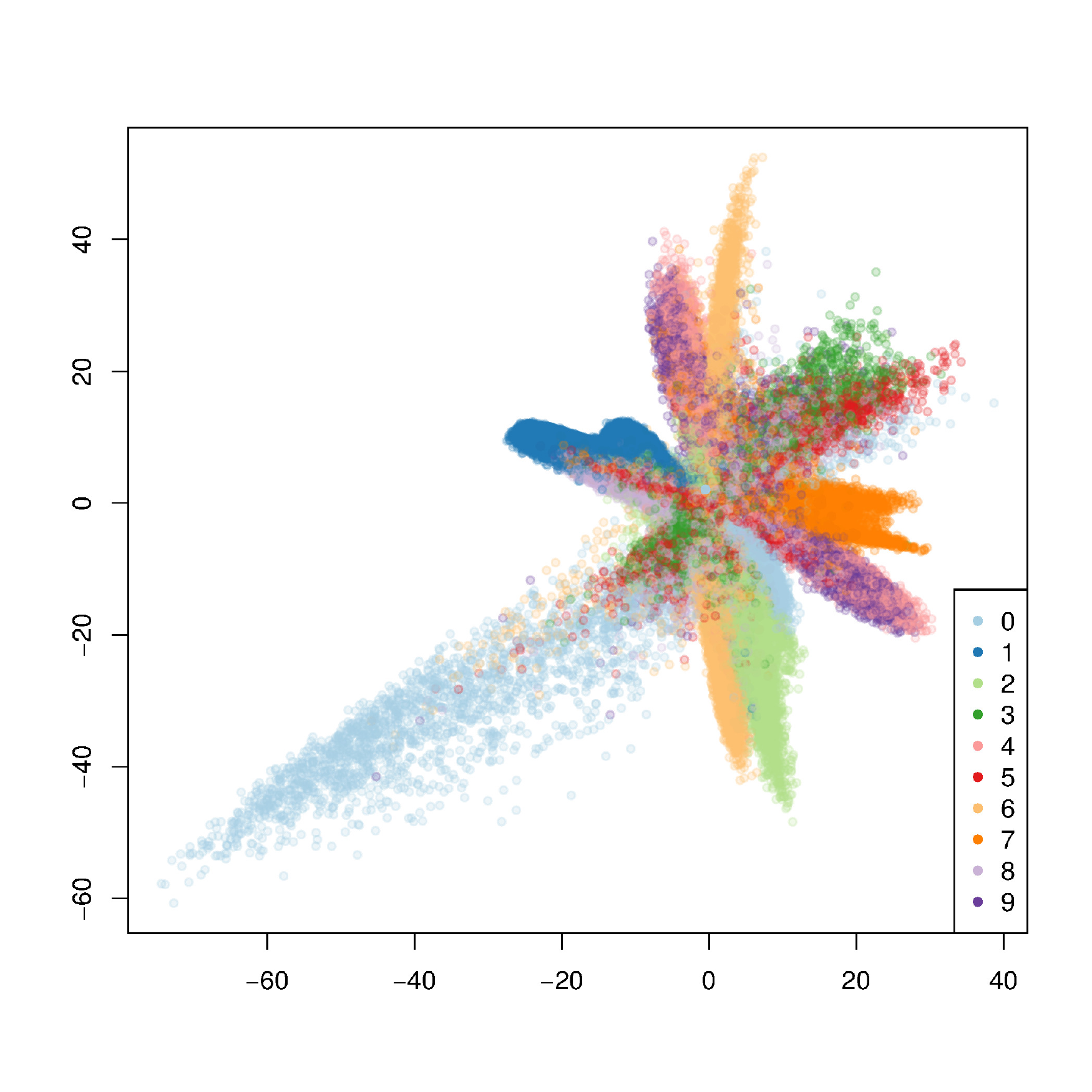}
\caption{Dimensionality reduced representations of digits for MNIST for autoencoder linear model tree.}
\label{fig:dimreduced-linear}
\end{figure}

\begin{figure}[t]
\centering
\includegraphics[width=0.45\textwidth]{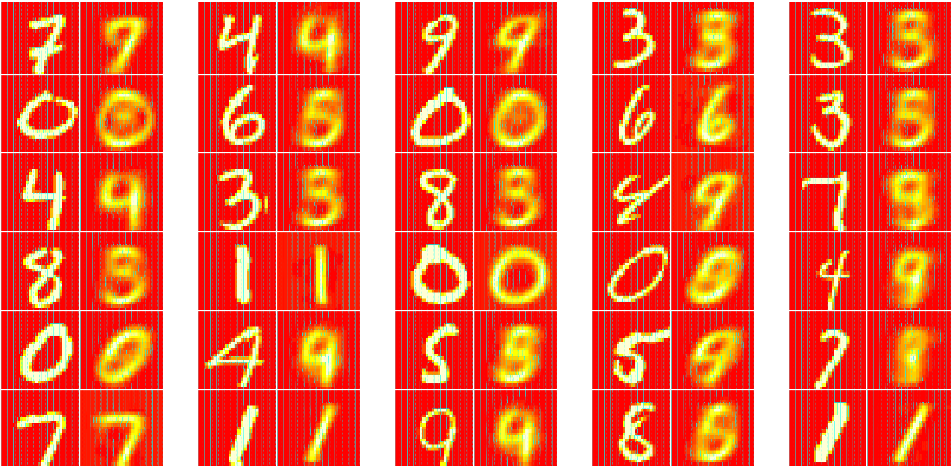}
\caption{A sample of original (left columns) and reconstructed (right columns) images. Selection is random. Reconstructed from 2d reduction with autoencoder
linear model trees.}
\label{fig:mnist-reconstr-linear}
\end{figure}

We experiment with this extension on the MNIST dataset. Figure 8 shows the resulting two- dimensional representations for different digits where we see that classes are very well-separated even in two dimensions, indicating that the model has effectively captured the underlying distribution.  In contrast to Figure 3, autoencoder linear model trees provide a much smoother distribution in the hidden representation space, and move away from the clustering-like behavior to a degree. This validates our intuition that linear models would help incorporate a distributed representation in addition to locality. We see in Figure 2(left) that with such trees we can get smaller reconstuction error on MNIST data, as also observed on a sample of reconstructed images in Figure 9. 

We experiment with this extension on the MNIST dataset. Figure~\ref{fig:dimreduced-linear} shows the resulting two-dimensional representations for different digits
where we see that classes are very well-separated even in two dimensions, indicating that the model has effectively captured the underlying distribution. In contrast to Figure~\ref{fig:dimreduced}, autoencoder linear model trees provide a much smoother distribution in the hidden representation space, and move away from the clustering-like behavior to a degree. This validates our intuition that linear models would help incorporate a distributed representation in addition to locality.
We see in Figure~2 (left) that with such trees we can get smaller reconstuction error on MNIST data, as also observed on a sample of reconstructed images in Figure~9.

\section{Conclusions and Discussion}

We discuss an autoencoder model with soft decision trees as encoder and decoder. We apply our model in a dimensionality reduction setting. The model is shown to have comparable or better reconstructive power than autoencoder perceptrons, when reducing to a small number of dimensions. Autoencoder trees provide a hierarchical decomposition in the input space and the space of the hidden representation, by making both the encoding and decoding hierarchical.

Autoencoder trees can be conceptualized as doing a soft hierarchical clustering on the data, and doing a dimensionality reduction within the clusters. As opposed to applying a hierarchical clustering algorithm and then reconstructing each cluster by its centroid, autoencoder trees provide dimensionality reduction coupled with a clustering-like behavior. Furthermore, decoding process is also hierarchical in autoencoder trees. The use of a soft convex combination of leaves implies that an instance can be efficiently modeled as a combination of multiple leaf nodes, allowing to model things such as a background factor that is added to many instances.

The hidden representation learned by an autoencoder tree can be fed to a supervised learner for classification or regression. We see that though the training is unsupervised, the autoencoder tree finds internal nodes or leaves that become increasingly responsive to single or few classes. The combination of gating values from the root to a node defines the boundaries of a locality and hidden representations learned at different levels can be considered as representations at different resolutions or granularities. Feeding these representations at multiple resolutions may improve prediction performance in a supervised setting; this will be an interesting future research direction. Another promising avenue for future work would be to incorporate recently proposed budding trees \cite{irsoy2014budding} which can learn the structure of the tree adaptively from data.


\bibliographystyle{ieeetr}
\bibliography{ref}

\end{document}